\documentclass[10pt, conference]{IEEEtran}

\usepackage{multicol}
\usepackage[bookmarks=true]{hyperref}
\usepackage{amsmath}
\usepackage{framed}
\usepackage{amssymb}
\usepackage{bm}
\usepackage[version-1-compatibility]{siunitx}
\usepackage{arydshln}

\usepackage{booktabs}
\usepackage{amsthm}
\usepackage{graphicx}
\usepackage[dvipsnames, table]{xcolor}
\usepackage{tikz}
\usetikzlibrary{positioning}
\usetikzlibrary{calc}
\usetikzlibrary{decorations.pathmorphing,patterns}
\usepackage{blindtext}
\usepackage{amsmath}

\begin{document}

\title{Convex Controller Synthesis for Robot Contact}

\author{
\IEEEauthorblockN{Hung Pham\IEEEauthorrefmark{1}, Quang-Cuong
  Pham\IEEEauthorrefmark{1}\IEEEauthorrefmark{2}} 
\IEEEauthorblockA{\IEEEauthorrefmark{1}Eureka Robotics, Singapore. Email:
  \texttt{hungpham@eurekarobotics.com}}
\IEEEauthorblockA{\IEEEauthorrefmark{2}School of Mechanical and
  Aerospace Engineering, Nanyang Technological University, Singapore}
}


\maketitle

\begin{abstract}
  Controlling contacts is truly challenging, and this has been a major
  hurdle to deploying industrial robots into
  unstructured/human-centric environments. More specifically, the main
  challenges are: (i) how to ensure stability at all times; (ii) how
  to satisfy task-specific performance specifications; (iii) how to
  achieve (i) and (ii) under environment uncertainty, robot parameters
  uncertainty, sensor and actuator time delays, external
  perturbations, etc. Here, we propose a new approach -- Convex
  Controller Synthesis (CCS) -- to tackle the above challenges based
  on robust control theory and convex optimization. In two physical
  interaction tasks -- robot hand guiding and sliding on surfaces with
  different and unknown stiffnesses -- we show that CCS controllers
  outperform their classical counterparts in an essential way.
\end{abstract}

\IEEEpeerreviewmaketitle

\section{Introduction}
\label{sec:introduction}

Controlling the behavior of an industrial robot when it comes to
contact with an unknown environment is truly challenging. To date,
contact tasks for industrial, position-controlled, robots, such as
assembly, deburring, or polishing, have always required a highly
accurate model (geometry, stiffness, friction coefficient) of the
environment. In particular, there are very few, if any,
production-deployed instances of industrial robots physically
interacting with environments whose stiffnesses are unknown. More
specifically, the challenges are threefold:


\begin{enumerate}
\item How to ensure the \emph{stability} of the robot at all times:
  instability may lead to catastrophic consequences such as excessive
  contact forces that may damage the robot, the workpiece or, at
  worst, harm the human operator;
\item How to satisfy task-specific performance specifications, which
  may include minimizing force/position tracking errors, fast
  response, noise attenuation, disturbance rejection;
\item How to achieve (1) and (2) under environment uncertainty, robot
  parameters uncertainty, sensor and actuator time delays, external
  perturbations, etc.
\end{enumerate}

\begin{figure}[tp]
  \tikzstyle{textcube} = [color=black, fill=white]
  \centering
  \begin{tikzpicture}
    \node[anchor=south west,inner sep=0] (image) at (0,0)
    {\includegraphics[width=0.33\textwidth]{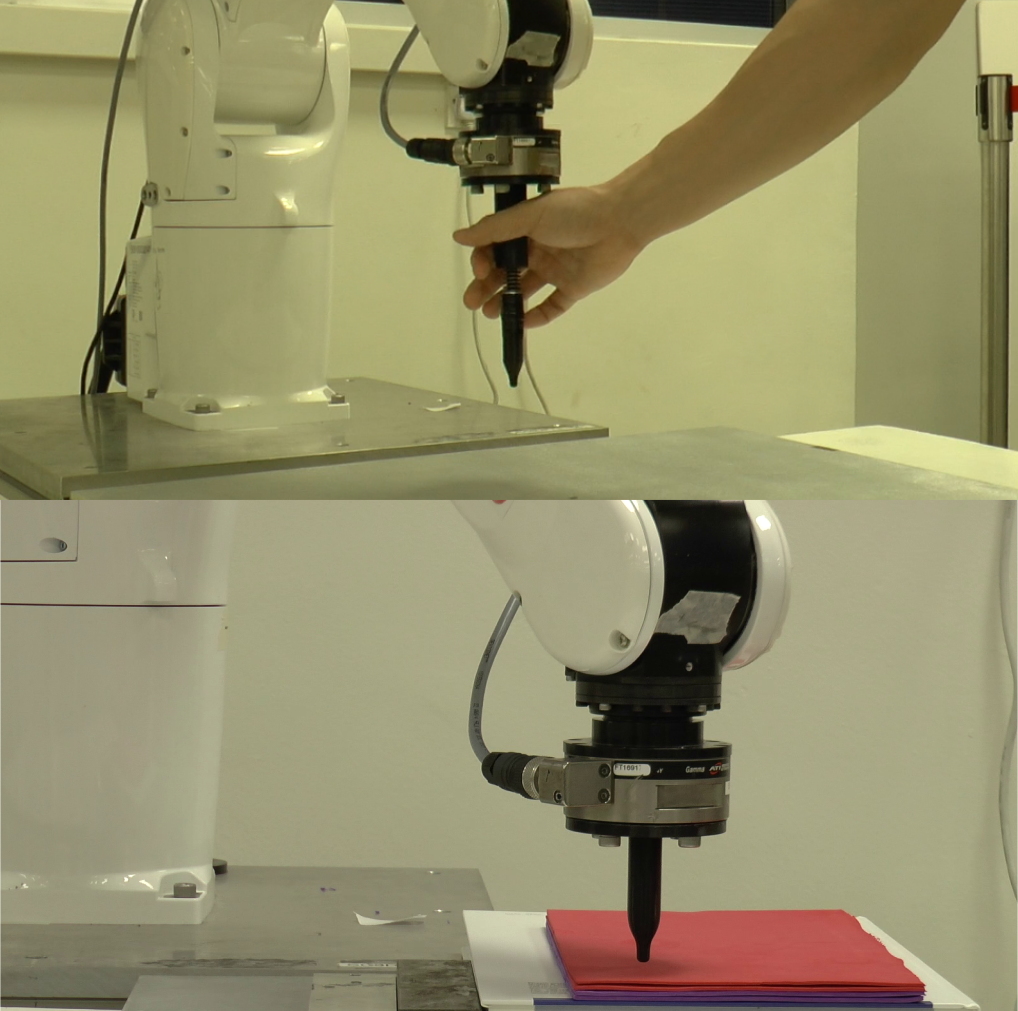}};
    \begin{scope}[x={(image.south east)},y={(image.north west)}]




      


      \node[textcube, anchor=north west, draw] at (0, 0.5) {Sliding on different
      surfaces};
      \node[textcube, anchor=north west, draw] at (0, 1.0) {Robot hand guiding};

    \end{scope}
  \end{tikzpicture}
  \caption{\label{fig:overview} Experimental setup. Top: Robot hand
    guiding (Section~\ref{sec:experiment}). Bottom: Sliding on
    multiple surfaces with different and unknown stiffnesses
    (Section~\ref{sec:exper-eval-force}). The video of the experiments
    is available at \url{https://youtu.be/uqYXVB5Sqlg}.}
\end{figure}

There has been substantial work on contact controllers that can deal
with environment uncertainty, in particular, unknown environment
stiffness. One approach consists in estimating the environment
stiffness in real time and adapting controller gains
accordingly~\cite{roy2002adaptive, kroger2004adaptive,
  stolt2012adaptation, rossi2016implicit, polverini2017data}. One
major limitation is the comparatively low sensitivity and speed of the
stiffness estimator, which, in turn, severely restricts the reactivity
of the controller. Other approaches are based on robust control
theory~\cite{polverini2017implicit, polverini2017robust} or
Model-Predictive Control~\cite{polverini2016performance}, but so far
such approaches have been restricted to simple robot/environment
models and limited ranges of environment stiffnesses (about two
times).



In this paper, we propose a new approach -- Convex Controller
Synthesis (CCS) -- to tackle the above challenges. Our approach relies
on robust control theory as a \emph{systematic} modeling framework,
and numerical convex optimization as a synthesis tool. Unlike
approaches based on stiffness estimation, there is no need here to
estimate environment stiffness nor to change controller gains, which
enables fast and reactive control. Compared to previous approaches
based on robust control or MPC, our systematic framework can model
most of the relevant sources of uncertainties (robot parameter
uncertainties, sensor and control time delays, external
perturbations), while handling a large range of environment
stiffnesses (up to 27 times, as shown in the experiments). A more
detailed discussion of related work is offered in
Section~\ref{sec:related-work}.

Specifically, our contributions are:
\begin{itemize}
\item we formulate contact control problems (including
  Admittance/Impedance Control and Direct Force Control\,\footnote{In
    some articles, Admittance Control is used to \emph{indirectly}
    regulate the contact force by modulating the robot's
    admittance. Therefore, we use the term ``Direct'' here to mean
    that the objective is to \emph{directly} track a desired contact
    force.}), and relevant sources of uncertainties in the framework
  of robust control theory;
\item we numerically address that formulation based on appropriate
  tools (Q-parameterization and convex optimization);
\item we demonstrate, in two physical experiments -- robot hand
  guiding and robot sliding on surfaces with different and unknown
  stiffnesses -- that CCS controllers outperform their classical
  counterparts in an essential way.
\end{itemize}

Note that our experiments are performed with position-controlled
industrial robots, which involve significantly more difficulties when
it comes to contact control (see Section~\ref{sec:an-example:-force}
for more detail) as compared to torque-controlled robots. The results
are therefore widely applicable, as the overwhelming majority of
robots in the industry are position-controlled, owing to their high
precision and cost-effectiveness~\cite{suarez2018can}.

The paper is organized as follows. In
Section~\ref{sec:core-framework}, we present the core framework, which
relies on appropriately selected and contextualized elements of robust
control theory and convex optimization. In
Section~\ref{sec:lead-through-with}, we delve into the synthesis of
convex controllers for contact. In Section~\ref{sec:experiments}, we
report the results of two physical contact experiments: robot hand
guiding and sliding on multiple surfaces. Finally, we discuss the
significance of the experimental results and conclude by sketching
some future research directions (Section~\ref{sec:conclusion}).

\section{Related work}
\label{sec:related-work}

{\small

  As mentioned, one approach to dealing with uncertainties in
  environment stiffness is to estimate the stiffness online and adapt
  the controller gains accordingly. In~\cite{roy2002adaptive,
    kroger2004adaptive, stolt2012adaptation, rossi2016implicit},
  researchers used a Least-Square-based estimator to estimate the
  stiffness of the environment, which is then used to select the
  actual gains of the force control loop. More recently,
  \cite{polverini2017data} proposed to use Virtual Reference Feedback
  Tuning~\cite{campi2002virtual} to adapt the controller directly to
  stiffness measurements. Relevant to this approach, a recent work
  \cite{roveda2017} employs a two-controllers architecture where one
  layer adapts to the unknown external environment.  These approaches
  work well when the control objective is simple, such as tracking a
  constant contact force throughout the task. However, for tasks that
  involve switching among multiple modalities (e.g., rapidly making
  and breaking contacts with different surfaces), stiffness estimation
  is less effective, making it harder to select appropriate control
  gains.

  In recent years, there has been a shift in force control research
  toward designing \emph{robust} controllers that are stable across a
  range of environments. This also reduces the need for \emph{online
    estimation}, paving the way for more difficult assembly or
  interaction problems. In~\cite{polverini2017implicit,
    polverini2017robust}, set-invariance theory is used to handle
  \emph{moderate} level of uncertainty in environment stiffness (about
  two times, which is much less than what CCS can achieve).
  Model-Predictive Control is another common approach in control
  engineering~\cite{mayne2000constrained} that improves the robustness
  of the system: this was applied to force control
  in~\cite{polverini2016performance}. The two approaches just
  mentioned assume a relatively simple model of the environment/robot
  (elastic environment with double-integrator robot dynamics). Other
  relevant types of uncertainties, such as time-delays or
  time-discretization, have not been considered.

  The theory of passive systems leads to yet more approaches to the
  design of robust force controllers. The central observation is that
  a combination of passive systems is passive, and therefore
  stable~\cite{slotine1991applied}. Accordingly, a line of research
  consists in developing control algorithms that make the robot
  dynamics passive~\cite{balachandran2017passivity,
    albu2007unified}. While this approach works well in the reported
  experiments, its main drawback is that passivity is a very
  conservative property: a system can be stable without being
  passive. In particular, time-delays and time-discretization in the
  robot control loop lead to extremely restrictive passivity
  specifications, which are detrimental to performance. Thus,
  passivity-based controllers are stable but not necessarily
  high-performing: it is difficult to achieve the kind of performance
  specifications described in the present paper.

  Designing robust control laws has been an active research topic
  since the 80's, starting from the seminal
  result~\cite{youla1976modern}.  Important results on
  $\mathcal{H}_{\infty}/\mathcal{H}_{2}$ were derived by
  \cite{doyle1983synthesis} and later developed into a powerful
  theory~\cite{doyle2013feedback,zhou1996robust}.  The connection to
  convex optimization was initially explored in~\cite{boyd1990linear}
  and more recently in~\cite{wang2016system}.  The present paper
  examines and applies results in robust control and convex
  optimization to the handling of contact.

}

\section{General Control Problem Formulation and Convex Controller
  Synthesis (CCS)}
\label{sec:core-framework}

\subsection{General Control Problem Formulation}
\label{sec:problem-formulation}

We model contact dynamics by a discrete-time linear and time-invariant
(LTI) system (a justification for this modeling choice is given when
we discuss an actual control system in
Section~\ref{sec:an-example:-force}):
\begin{equation}
  \label{eq:2}
  \begin{aligned}
    \mathbf{x}[n+1] &=&  A    &\mathbf{x}[n]& + B_{1}  &\mathbf{w}[n]& + B_{2} \mathbf{u}[n], \\
    \mathbf{z}[n] &=&   C_{1} &\mathbf{x}[n]& + D_{11} &\mathbf{w}[n]& + D_{12} \mathbf{u}[n], \\
    \mathbf{y}[n] &=&   C_{2} &\mathbf{x}[n]& + D_{21} &\mathbf{w}[n]& + D_{22} \mathbf{u}[n].
  \end{aligned}
\end{equation}
Here, $\mathbf{x}[n]$ is the vector of internal states at time $n$;
$\mathbf {u}[n], \mathbf {w}[n]$ are respectively the control inputs
and the exogenous inputs; $\mathbf {y}[n], \mathbf {z}[n]$ are
respectively the measured outputs and the exogenous outputs. Note that
actuator/sensor time-delays can be easily modeled by introducing
additional states.

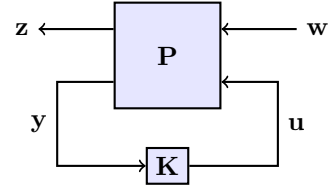
\begin{figure}[ht]
  \centering
  \tikzstyle{block} = [draw,fill=blue!10,minimum size=1em, thick]
  \begin{tikzpicture}
    \node[block, minimum size=4em] (x) {$\mathbf {P}$};
    \node[block, below=0.5cm of x] (K) {$\mathbf K$};
    \draw[->, thick] ([yshift=10pt]x.west) -- node[anchor=east, pos=1.0] {$\mathbf {z}$} +(-1cm,0pt);
    \draw[<-, thick] ([yshift=10pt]x.east) -- node[anchor=west, pos=1.0] {$\mathbf w$} +(+1cm,0pt);
    \draw[->, thick] ([yshift=-10pt]x.west) -- +(-0.75cm,0pt) |- node[pos=0.25, anchor=east] {$\mathbf y$} (K);
    \draw[<-, thick] ([yshift=-10pt]x.east) -- +(+0.75cm,0pt) |- node[pos=0.25, anchor=west] {$\mathbf u$} (K);
  \end{tikzpicture}
  \caption{\label{fig:lft} General Control Problem Formulation. The
    plant $\mathbf{P}$ maps exogenous inputs $\mathbf w$ and control
    inputs $\mathbf u$ to exogenous outputs $\mathbf z$ and measured
    outputs $\mathbf y$. Note that the measured outputs $\mathbf y$
    are inputs to the controller $\mathbf{K}$, while the control
    inputs $\mathbf u$ are its outputs.}
\end{figure}

Denote the Z-transform of a signal by a single bold-faced letter
(e.g. $\mathbf x$). Taking the Z-transform of both sides of
equation~\eqref{eq:2} and assuming zero initial conditions yield the
following transfer matrix representation:
\begin{equation}
  \label{eq:1}
  \left[
  \begin{array}{c}
    \mathbf{z} \\ \hdashline[2pt/2pt]  \mathbf y
  \end{array}
  \right]
  =
  \begin{bmatrix}
    \mathbf {P}_{11}(z) & \mathbf {P}_{12}(z) \\
    \mathbf {P}_{21}(z) & \mathbf {P}_{22}(z)
  \end{bmatrix}
  \left[
  \begin{array}{c}
    \mathbf{w} \\ \hdashline[2pt/2pt]  \mathbf u
  \end{array}
  \right]
\end{equation}
where $\mathbf P_{ij}(z):= C_{i}(Iz - A)^{-1}B_{j} + D_{ij}$.  This
representation is more convenient during modeling, while the
state-space representation offers efficiency in simulation and
analysis.

We consider controllers that are also discrete-time LTI systems:
\begin{equation}
  \label{eq:9}
  \begin{aligned}
    \mathbf{x}^{(K)}[n+1] &=&  A^{(K)}    &\mathbf{x}^{(K)}[n]&  + B^{(K)} \mathbf{y}[n], \\
    \mathbf{u}[n] &=&   C^{(K)} &\mathbf{x}^{(K)}[n]&  + D^{(K)} \mathbf{y}[n],
  \end{aligned} 
\end{equation}
here the superscript $\Box^{(K)}$ denotes quantities internal to the
controller.  Note that the measured outputs $\mathbf y[n]$ are inputs to
the controller, while the control inputs $\mathbf u[n]$ are its
outputs, see Fig.~\ref{fig:lft} for an illustration. The Z-transform
of the controller is also a transfer matrix:
\begin{equation}
  \label{eq:5}
  \mathbf K(z) = C^{(K)}(Iz - A^{(K)})^{-1}B^{(K)} + D^{(K)}.
\end{equation}

Suppose that controller $\mathbf K(z)$ stabilizes a given plant
$\mathbf P(z)$, the closed-loop system dynamics is an LTI system that
maps the exogenous inputs $\mathbf w$ to the exogenous outputs
$\mathbf z$. The closed-loop transfer matrix $\mathbf H(z)$ is given
by~\footnote{A subtle issue in the design of the controller is the
  well-posedness of the feedback loop, which guarantees the existence
  of $\mathbf H(z)$.  A necessary and sufficient condition is that the
  matrix
\begin{equation}
  \label{eq:11}
  I - \mathbf P_{22}(\infty) \mathbf K(\infty) = I - D_{22}D_{k}
\end{equation}
is invertible~\cite{zhou1996robust}. In the context of robotic
applications, $\mathbf P$ represents a physical system, which often
has zero feed-through $D_{22}=0$ due to time-delays, hence, satisfying
the well-posedness condition trivially.
}:
\begin{equation}
  \label{eq:7}
  \begin{aligned}
    \mathbf{z} &= (\mathbf P_{11}(z)  + \mathbf P_{12}(z) \mathbf K(z)  (I - \mathbf P_{22}(z) \mathbf K(z) )^{-1}\mathbf P_{21}(z) ) \mathbf w \\
    &=: \mathbf H(z) \mathbf w.
  \end{aligned}
\end{equation}

The controller synthesis problem is to find a controller
$\mathbf K(z)$ such that the closed-loop system is stable and that the
closed-loop transfer matrix $\mathbf H(z)$ achieves the desired
specifications of the given task, which are specified via the
exogenous inputs $\mathbf w$ and outputs $\mathbf z$.


\subsection{Example of General Control Problem Formulation: Direct
  Force Control for a position-controlled robot}
\label{sec:an-example:-force}

To illustrate the General Control Problem Formulation, we show here
how to cast a classical problem in industrial robotics, Direct Force
Control for a position-controlled robot (see
e.g.~\cite{suarez2018can}), into that formalism.

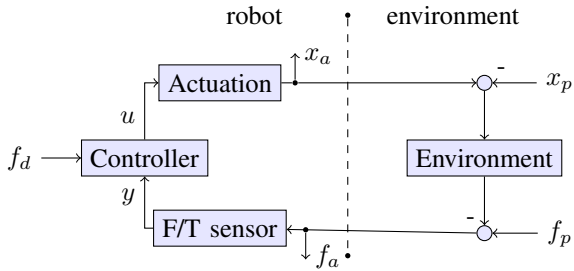
\begin{figure}[ht]
  \centering
  \tikzstyle{block} = [draw,fill=blue!10,minimum size=1em]
  \tikzstyle{sum} = [draw, fill=blue!10, circle, inner sep=2pt]
  \tikzstyle{junction} = [fill, circle, inner sep=0pt, minimum size=2pt]

  \begin{tikzpicture}

    \node[block] (R) at (0, 0) {Actuation};
    \node[left of=R] (j1) {};
    \node[block, below of=j1] (K) {Controller};
    \node[block, below= 41pt of R] (F) {F/T sensor};

    \node[junction, right of=R] (j2) {};
    \node[block, right= 75pt of K] (E) {Environment};
    \node[sum, above of=E] (sum-top-of-E) {};
    \node[sum, below of= E] (s1) {};

    \node[left=15pt of K] (fd){$f_{d}$};

    \node[right of=sum-top-of-E] (xp) {$x_{p}$};
    \node[right of=s1] (fp) {$f_{p}$};
    \node[left=20pt of j1] (u) {};

    \draw[->] (xp) -- node[pos=0.8, anchor=south] {-} (sum-top-of-E);
    \draw[->] (E) -- node[pos=0.9, anchor=east] {-}  (s1);
    \draw[->] (fp) -- (s1);
    \draw[->] (s1) -- node[junction, pos=0.9] (junction FT) {} (F);
    \draw[->] (R) -- (j2) -- (sum-top-of-E);
    \draw[->] (sum-top-of-E) -- (E);
    \draw[->] (K) |- node[pos=0.2, anchor=east] {$u$} (R);
    \draw[->] (F) -| node[pos=0.8, anchor=east] {$y$}(K);
    \draw[->] (fd) -- (K);

    \draw[->] (j2) -- node[pos=0.9, anchor=west] {$x_{a}$} ([yshift=10pt] j2.north);
    \draw[->] (junction FT) -- node[pos=0.9,anchor=west] {$f_{a}$} ([yshift=-10pt] junction FT.south);

    \node[junction] (sep1) at (1.7, 0.9) {};
    \node[junction] (sep2) at (1.7, -2.3) {};
    \draw[dashed] (sep1) -- (sep2);
    \node[left=20pt of sep1, anchor=east] {robot};
    \node[right=10pt of sep1, anchor=west] {environment};
    
  \end{tikzpicture}

  \caption{\label{fig:force-control} Force Control for a
    position-controlled robot, formulated as a General Control
    Problem. The exogenous inputs are the desired contact force $f_d$,
    the perturbation $x_p$ in the robot position, and the perturbation
    $f_p$ in the contact force. The control input is the position
    command $u$ to the robot actuators. The exogenous outputs are
    actual robot position $x_a$ and the actual contact force
    $f_a$. The measured output is the measured contact force $y$.}
\end{figure}

Most industrial robots are position-controlled, i.e. the user
specifies a desired position $u$, and the robot internal controller
$R(z)$ -- on which the user usually has no authority -- tries to
achieve that desired position as precisely as possible using a
high-gain loop. To perform Direct Force Control -- i.e. tracking a
desired contact force $f_d$ -- the idea is to obtain a measurement $y$
of the contact force through a Force/Torque (F/T) sensor mounted at
the robot flange, and to compute an appropriate position command
$u$. This scheme is illustrated in Fig.~\ref{fig:force-control}.
 
Note at this point that the LTI formulation is well justified: the
environment can be appropriately modeled as a linear spring-damper
system, while the input/output relationships of the actuation and of
the F/T sensor can be appropriately modeled as linear filters with
time-delays.

To cast this scheme into the General Control Problem Formulation, one
may define the following signals:
\begin{itemize}
\item the exogenous inputs as the desired contact force $f_d$, the
  perturbation $x_p$ in the robot position, and the perturbation $f_p$
  in the contact force;
\item the control input is the position command $u$ to the
  robot;
\item the exogenous outputs are the actual robot position $x_a$ and
  the actual contact force $f_a$;
\item the measured output is the measured contact force $y$ from the
  F/T sensor.
\end{itemize}
This yields the following open-loop plant transfer function and the
corresponding closed-loop system transfer function:
\begin{equation}
  \label{eq:13}
  \left[
  \begin{array}{c}
    x_{a} \\ f_{a} \\ \hdashline[2pt/2pt] y \\ f_{d}
  \end{array}
  \right]
  = \mathbf P(z) 
  \left[
  \begin{array}{c}
    f_{p} \\ x_{p} \\ f_{d} \\ \hdashline[2pt/2pt]  u
  \end{array}
  \right],\ \
  \begin{bmatrix}
    x_a \\ f_a
  \end{bmatrix} = \mathbf H(z)
  \begin{bmatrix}
    f_p \\ x_p \\ f_d
  \end{bmatrix}.
\end{equation}

Besides guaranteeing close-loop stability, one can enforce performance
specifications such as:
\begin{itemize}
\item the robot should maintain a stable contact with the environment,
  which can be time-varying;
\item the robot should track a step reference force signal without
  steady-state error, and with a sufficiently high bandwidth;
\item attenuation of noises from the sensors and motors.
\end{itemize}

These three performance specifications correspond in fact to three
elements of the transfer matrix $\mathbf H(z)$.  Therefore, by
appropriately constraining and optimizing these elements, one can
achieve the stated specifications.

\subsection{Q-parameterization and CCS}
\label{sec:q-param-stable}

A controller is said to be \emph{stabilizing} if the closed-loop
system is stable. The set of all closed-loop transfer matrices
$\mathbf H(z)$ achievable by stabilizing controllers
\begin{equation}
  \label{eq:26}
  \mathcal {H} = \{\mathbf H(z) \mid \mathbf K(z) \text{ is stabilizing and satisfies} \eqref{eq:7}\}
\end{equation}
has in fact a very simple structure: it can be parameterized
affinely~\cite{youla1976modern}. Specifically, there exist three
transfer matrices $\mathbf {T}_{1}(z), \mathbf T_{2}(z), \mathbf
T_{3}(z)$ such that for any $\mathbf H(z) \in \mathcal{H}$, there is a
stable transfer matrix $\mathbf Q(z)$ such that
\begin{equation}
  \label{eq:12}
  \mathbf H(z) = \mathbf T_{1}(z) + \mathbf T_{2}(z) \mathbf Q(z) \mathbf T_{3}(z).
\end{equation}
Conversely, for any stable transfer matrix $\mathbf Q(z)$, the
transfer matrix $\mathbf H(z)$ defined by Eq.~\eqref{eq:7} is a
valid closed-loop transfer matrix that is realized by a stabilizing
controller.

Specializing to stable open-loop plants, the coefficients
$\mathbf {T}_{1}(z), \mathbf T_{2}(z), \mathbf T_{3}(z)$ are
relatively simple~\cite{zhou1996robust}:
\begin{equation}
  \mathbf T_{1}(z) = \mathbf P_{11}(z), \mathbf T_{2}(z) = \mathbf P_{12}(z), \mathbf T_{3}(z) = \mathbf P_{21}(z) .
\end{equation}
The controller can be recovered from $\mathbf Q(z)$ using the
following relation:
\begin{equation}
  \label{eq:4}
  \mathbf K(z) = (I + \mathbf Q(z) \mathbf P_{22}(z)^{-1}) \mathbf Q(z).
\end{equation}
While it is possible to use the above equation to explicitly compute
the controller, it is not recommended.  Rather, one can implement a
controller $\mathbf K(z)$ directly by constructing a feedback loop of
$\mathbf Q(z)$ and $\mathbf P_{22}(z)$.

Note that $\mathcal H$ is an affine set: for any two closed-loop
transfer matrices $\mathbf H_{1}, \mathbf H_{2} \in \mathcal{H}$, one
can obtain a one-parameter family of closed-loop transfer matrix:
\begin{equation}
  \label{eq:14}
  \alpha \mathbf H_{1}(z) + (1 - \alpha) \mathbf H_{2}(z) \in \mathcal H, \alpha \in R.
\end{equation}
This follows easily from the linearity of $\mathbf Q$ in the
expression of $\mathbf H(z)$ Eq.~\eqref{eq:12}, and the observation
that linear combinations of stable transfer matrices are stable.
Since affine sets can be handled efficiently, this property is perhaps
the most fundamental to our numerical synthesis of controllers.

As a result, one can formulate a general Convex Controller Synthesis
problem as a convex optimization problem:
\begin{equation}
  \label{eq:10}
  \begin{aligned}
    \mathrm{argmin}_{\mathbf Q(z) \text{stable}} \quad& f(\mathbf H(z)) \\ \textrm{subject
      to}\quad& \mathbf H(z) \in \mathcal {H} \\ & \mathbf H(z) \in
    \mathcal C_{i}, i =0, \dots, N_{c}, 
  \end{aligned}
\end{equation}
where $f$ is a convex objective function and the $\mathcal C_{i}$'s
are convex sets arising from performance specifications.

\subsection{Synthesis by numerical optimization}
\label{sec:contr-synth-as}

To obtain a finite-dimensional approximation, we follow the
computational approach proposed in~\cite{boyd1990linear}. In
particular, we select a set of basis stable transfer matrices
$\{\mathbf Q_{i}, i=0, \dots, n-1\}$, and approximate the optimal
$\mathbf Q(z)$ by
$\mathbf Q(z) = \sum_{i=0}^{n-1}\theta_{i} \mathbf Q_{i}(z),$ where
$\theta_{i}$ are real parameters. The closed-loop transfer matrix
$\mathbf H(z)$ is next given by:
\begin{equation}
  \label{eq:15}
  \mathbf H(z) = \mathbf T_{1}(z) + \sum_{i}^{n-1} \theta_{i} 
  \mathbf T_{2}(z) \mathbf Q_{i}(z) \mathbf T_{3}(z).
\end{equation}
Note that $\mathbf H(e^{j\omega T_{s}})$, the frequency response at
angular velocity $\omega$, is linear in the parameter vector
$\bm \Theta := [\theta_{0},\dots,\theta_{n-1}]^{\top}$:
\begin{equation}
  \label{eq:27}
  \mathbf H(e^{j\omega T_{s}}) =  \mathbf T_{1}(e^{j\omega T_{s}}) + 
  \bm{\mathcal{T}}(e^{j\omega T_{s}}) \bm{\Theta},
\end{equation}
where $\bm {\mathcal{T}}(e^{j\omega T_{s}})$ is a complex-valued block
matrix, obtained by appropriately rearranging Eq.~\eqref{eq:15}.

By the linearity of the inverse Z-transform, the closed-loop impulse
response is also linear in the parameter vector:
\begin{equation}
  \label{eq:28}
  \mathbf H[n] = \mathbf T_{1}[n] + \bm{\mathcal{T}}[n] \bm{\Theta}.
\end{equation}

By expressing performance specifications as convex constraints and
convex objective functions on the frequency response and the impulse
response of the closed-loop transfer matrix, we obtain standard
numerical convex optimization problems, which can be solved
efficiently with standard convex optimization solvers.

In the sequel, we choose \emph{delayed unit-impulses} as basis
transfer functions. For a Single-Input Single-Output system with
scalar $\mathbf u$ and $\mathbf y$, this choice simplifies to
$Q_{i}(z) := z^{-i}$.  For general Multiple-Input Multiple-Output
system, there is a set of delayed impulses for each element of
$\mathbf Q$.  With this basis choice, computing the coefficients of
$\bm{\Theta}(e^{j\omega T_{s}})$ and $\bm {\Theta}[n]$ is
straightforward: each term of the sum in Eq.~\eqref{eq:15} is
simply a delayed transfer matrix of the previous one.

Some additional details on the computer implementation of CCS
controllers are given in the Supplementary Material:
\url{https://www.ntu.edu.sg/home/cuong/docs/CCS-sup.pdf}.

\section{CCS for contact tasks}
\label{sec:lead-through-with}

\subsection{Modeling contacts with unknown environments}
\label{sec:convex-formulation}

\subsubsection{Nominal and particular systems}



Following the practice of robust control theory, we distinguish two
types of systems: \emph{nominal} and \emph{particular}. The nominal
system captures the system dynamics in the \emph{expected operating
  condition}, while a particular system may display any dynamics
within a range of dynamic uncertainties.

We propose to synthesize controllers that achieve high performance
(optimal with respect to the given performance specifications) for the
nominal system and, at the same time, have stable closed-loop dynamics
for all particular systems. More concisely, we want to achieve
\emph{nominal performance and robust stability}.  Compared to the
strategy where the controllers try to achieve high performance at all
particular systems, the proposed strategy is less conservative, and
therefore can achieve better performance at and around the expected
operating condition.


\subsubsection{Modeling the environment}

\begin{figure}[ht]
  \tikzstyle{block} = [draw,fill=blue!10,minimum size=2em]
  \tikzstyle{sum} = [draw, fill=blue!10, circle, inner sep=2pt]
  \tikzstyle{junction} = [fill, circle, inner sep=0pt, minimum size=2pt]
  \tikzstyle{empty junction} = [inner sep=0pt, minimum size=0pt]
  \centering
  \begin{tikzpicture}

    \node[block, minimum size=3em] (R) at (0, 0) {$\mathbf R(z)$};
    \node[block, below of=R] (K) {${\mathbf K}$};
    \node[right=45pt of K, yshift=-3pt] (setpoint) {};
    \node[empty junction, right of=R, yshift=-8pt] (R right pt) {};

    \node[sum, above of=R, xshift=50pt] (sum) {};
    \node[sum, above of=R, xshift=-40pt] (sum-left) {};
    \node[above of=sum-left] (junction-aug-left) {};

    \node[junction, above of=R, xshift=40] (junction-aug) {};
    \node[block, above of=R, xshift=0pt] (env) {$\mathbf E(z)$};
    \node[block, above of=env, xshift=0pt] (env-aug) {$-\delta E_{\rm add}(z)$};

    \draw[->] (setpoint) -- node[pos=0., anchor=south] {setpoint}([yshift=-3pt] K.east);
    \draw[->] ([yshift=-8pt] R.east) -| node[pos=0.2, anchor=north] {$y$} (R right pt) |- ([yshift=+3pt] K.east);
    
    \draw[->] (K.west) -| ([xshift=-15pt, yshift=-8pt] R.west) -- node[pos=0.6, anchor=north] {$u$} ([yshift=-8pt] R.west) ;

    \draw[->, dashed] (junction-aug-left) -- node[pos=0.9,anchor=east] {-}
    node[pos=0.2, anchor=east] {$w_{E}$}
    (sum-left);

    \draw[->] (env-aug) --
    node[pos=0.9, anchor=south] {$z_{E}$}
    (junction-aug-left);

    \draw[->] ([yshift=8pt] R.east) -| node[pos=0.1, anchor=north] {$x$} (sum);
    \draw[->] (sum) -- (junction-aug) -- (env);
    \draw[->] (junction-aug) |- (env-aug);
    
    \draw[->] (env) -- (sum-left);
    \draw[->] (sum-left) |- node[pos=0.9, anchor=north] {$f$} ([yshift=8pt] R.west);

    \draw[->] ([xshift=20pt] sum.west) -- node[pos=0.0, anchor=west] {$x_{\rm env}$} node[pos=0.9, anchor=south]{-}(sum);

  \end{tikzpicture}
  \caption{\label{fig:inter} Interactions between the robot
    $\mathbf R(z)$, the controller $\mathbf K(z)$ and the environment
    $\mathbf E(z)$. Taking into account uncertainties amounts to
    ``closing the uncertainty loop'', making the dashed line $w_E$
    solid.}
\end{figure}
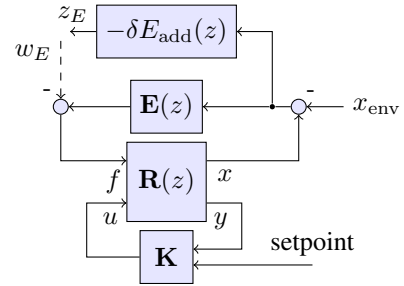

We model the unknown environment as a transfer function with
\emph{real additive parametric uncertainty}, characterized by an
unknown variable taking values in the interval $[0, 1]$. Here
$\delta=0$ corresponds to the nominal system, while $\delta=1$
corresponds to the worst-case particular system:
\begin{equation}
  \label{eq:3}
  \mathbf E_{p}(z) = \mathbf E_{n}(z) + \delta \mathbf E_{\rm add}(z), \quad \delta \in [0, 1].
\end{equation}
Such uncertainties can be incorporated in the overall system block
diagram as shown in Fig.~\ref{fig:inter}.  The terms in
Eq.~\eqref{eq:3} are linear transfer matrices, which can model
arbitrary linear systems such as springs, dampers, integrators, or any
linear combinations thereof.

Specifically, in our experiments, we consider two kinds of
uncertainties: a human operator with unknown stiffness interacting
with the robot (Experiment 1: robot hand guiding), and an environment
with unknown and varying stiffness (Experiment~2: sliding on different
surfaces).

The former can be modeled as
\begin{equation}
  \label{eq:29}
  \mathbf E = (k_{\rm hum} + b_{\rm hum} s) + \delta (\Delta_{k} + \Delta_{b} s),
\end{equation}
where $k_{\rm hum}, b_{\rm hum}$ are the nominal human stiffness and
damping, which apply when the operator is at rest. When the operator
exerts effort to interact with the robot, her muscles contract,
increasing the equivalent stiffness and damping
coefficients~\cite{Burdet2001}.

As for the environment with unknown and varying stiffness, it can be
modeled as
\begin{equation}
  \label{eq:30}
  \mathbf E = k_{\rm env} + \delta \Delta_{k}.
\end{equation}

\subsubsection{Modeling the robot}

We model the dynamics of the robot (industrial robot arm and F/T
sensor) as a transfer matrix $\mathbf R(z)$, mapping control input $u$
and actual force $f_{a}$ to actual distance moved $x$ and measured
force $f_{m}$:
\begin{equation*}
  \begin{bmatrix}
     x \\ f_{m}
  \end{bmatrix}
  =
  \begin{bmatrix}
    R_{11}(z) & R_{12} (z) \\ 
    R_{21}(z) & R_{22} (z)
  \end{bmatrix}
  \begin{bmatrix}
    f_a \\ u
  \end{bmatrix}.
\end{equation*}
The transfer function $\mathbf R(z)$ can either be derived from a
physical model, or experimentally identified. Effects such as
time-delays, which are difficult to handle in ``analytic'' approaches,
can be modeled directly in $\mathbf R(z)$ without difficulty. Other
robot architectures and effects can also be modeled, just to name a
few: torque-controlled or position-and velocity-controlled robot
control schemes; joint elasticity as well as effects of tip-mounted
(non-collocated) and joint-mounted (collocated) force sensors.




\subsection{Robust stability under real parametric uncertainty}
\label{sec:robust-stablity-real}

The key observation to guarantee robust stability is: for all values
of $\delta \in [0, 1]$, ``closing the uncertainty loop'' in
Fig.~\ref{fig:inter} must not destabilize the system. Assuming nominal
stability, by Nyquist's stability criterion~\cite{ogata2002modern},
the total open-loop gain of the uncertainty loop must not encircle the
$(-1,0)$ point in the complex plane in the clock-wise direction for
all values of $\delta$. Note that a passive system has an open-loop
gain that remains on the right-hand side of the $(-1,0)$ point and
therefore is stable (but also more conservative).

\begin{figure}[htp]
    \centering
  \begin{tikzpicture}
    \node[anchor=south west,inner sep=0] (image) at (0,0)
    {\includegraphics[]{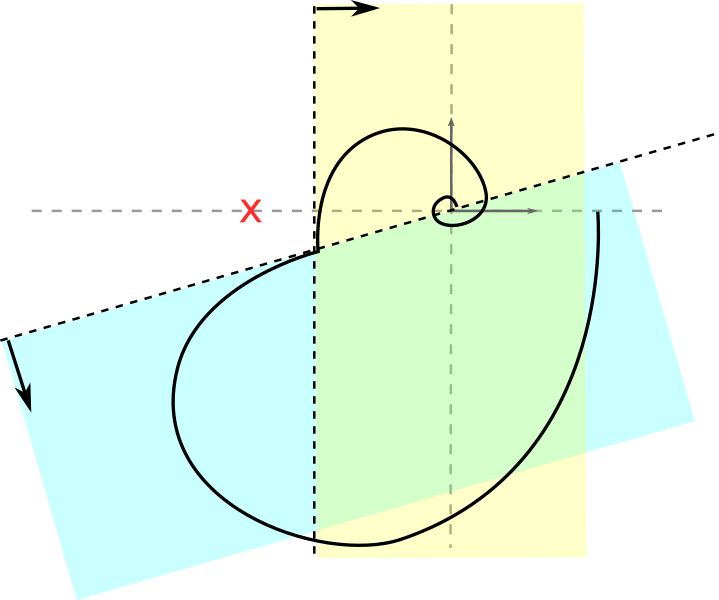}};
    \begin{scope}[x={(image.south east)},y={(image.north west)}]


      \node[anchor=east] at (0.35, 0.7) {$(-1, 0)$};

      \node[] (corner) at (0.45, 0.57) {};
      \node[] (corner annot) at (0.6, 0.4) {$\omega_{\rm corner}$};
      \node[] (HP1) at (0.1, 0.4) {$\mathcal{HP}_{1}$};
      \node[] (HP2) at (0.52, 0.92) {$\mathcal{HP}_{2}$};

      \draw[->, thick] (corner annot) -- (corner);
    \end{scope}
  \end{tikzpicture}
  \caption{\label{fig:Nyquist} The system is robustly stable if the
    Nyquist plot of the transfer function from $w_{E}$ to $z_{E}$ does
    not encircle the $(-1,0)$ point.}
\end{figure}
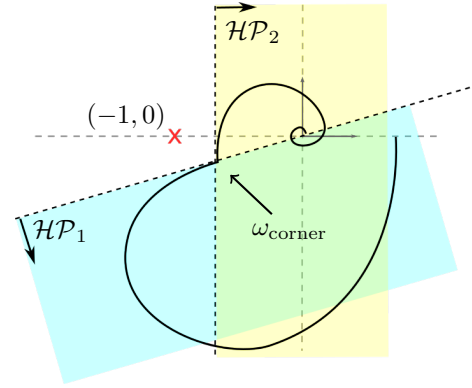

By making the two open ends of the uncertainty loop an exogenous input
and exogenous output, the loop gain is an element of the closed-loop
transfer matrix, which we denote by $\mathbf H_{\rm add}(z, \delta)$.
Therefore, in principle, one can ensure robust stability by enforcing
Nyquist's stability criterion in the controller synthesis procedure.

In its original form, however, Nyquist's stability constraint is
non-convex in the parameters $[\theta_{0},\dots,\theta_{n-1}]$. We
transform this constraint into a set of multiple convex
constraints. The main idea is to enclose different parts of the
Nyquist plot in different convex sets, which, together, enforce
Nyquist's stability criterion. A simple example is a pair of two
half-space constraints as shown in Fig.~\ref{fig:Nyquist}:
\begin{align}
  \label{eq:31}
  \mathbf H_{\rm add} (e ^{j\omega T_{s}}, \delta=1) \in \mathcal{HP}_{1}, & \quad \forall \omega \in [0, \omega_{\rm corner}]  \\
  \label{eq:31b}
  \mathbf H_{\rm add} (e ^{j\omega T_{s}}, \delta=1) \in \mathcal{HP}_{2}, & \quad \forall \omega \in [\omega_{\rm corner}, \omega_{\rm Nyquist}].
\end{align}
Since for $\delta=0$, $\mathbf H_{\rm add}(e ^{j\omega T_{s}},
\delta=0)$  satisfies both constraints in Eq.~\eqref{eq:31}
and~\eqref{eq:31b}, it follows from convexity that this holds for all
$\delta\in [0, 1]$, satisfying the robust stability condition.

Our proposed relaxation requires selecting a corner angular velocity
$\omega_{\rm corner}$ that the Nyquist plot for all velocities less
than this value lie the half-plane $\mathcal {HP}_{1}$ and for all
velocities greater lie in the second half-plane $\mathcal{HP}_{2}$. In
practice this is relatively easy to achieve.  Additionally, for robust
stability, it is common to limit the high-frequency spectrum of
dynamics to avoid exciting unmodeled dynamics.

\subsection{Convex formulation of some common performance
  specifications}
\label{sec:comm-perf-spec}

We now discuss some performance specifications commonly found in
robotic applications. We show that these specifications admit convex
formulations in the parameter space.

\subsubsection{Time-domain response}
\label{sec:time-doma-constr}

Specifications on time-domain responses concern the time-evolution of
the output signal in response to a given input signal. For example, in
force control, the contact force should have a first-order or
second-order critically damped response to a step input signal,
without overshoot. In robot hand guiding, the robot's motion in
response to a step input in the operator's ``desired position'' should
imitate the movement of an ideal mass-spring-damper system, so that it
is intuitive to the operator.

From Eq.~\eqref{eq:28}, one can see that the relationship between any
exogenous input/output pair is a linear function of the parameters
$[\theta_{0},\dots,\theta_{n-1}]$. Let $ij$ denote the indices of the
input/output signals of interest and $h_{ij}[n]$ denote the impulse
response, by concatenating the matrices, one obtains:
\begin{equation}
  \label{eq:32}
  h_{ij}[n] = 
  \begin{bmatrix}
    h_{ij}[0] \\
    \dots \\
    h_{ij}[N-1]
  \end{bmatrix} = \mathcal A_{ij} + \mathcal B_{ij} \bm{\Theta}.
\end{equation}
Here, $N$ is the horizon of the impulse responses, which is a
parameter to be chosen. Let $r_{j}[n]$ be the reference input signal
and $y_{i}^{(d)}[n]$ be the desired output signal. One can enforce
time-domain response shaping specifications by minimizing the cost
function:
\begin{equation}
  \label{eq:33}
  f(\bm{\Theta}):=\|y_{i}^{(d)}[n] - r_{j}[n] * \left \{\mathcal A_{ij} + \mathcal B_{ij}
   \bm {\Theta} \right \}\|_{2}.
\end{equation}
Note that any convex norm can be used.

Specifications such as no overshoot or minimum rise time can be
formulated as (convex) linear inequality constraints. The steady-state
value of a discrete-time transfer function is the sum of its impulse
response.

\subsubsection{Noise attenuation and passivity constraint}
\label{sec:phase-shaping}

One can attenuate the effect of noise with known frequency on certain
exogenous outputs by enforcing constraints on the frequency response
of the corresponding elements of the closed-loop transfer matrix
$\mathbf H(z)$.  In particular, to attenuate noise entering from the
$j$-th input on the $i$-th output, one may include the following
constraint:
\begin{equation}
  \label{eq:34}
  \|H_{ij}(e^{j\omega T_{s}})\| \leq w_{n}(\omega), \omega \in [\omega_{0}, \omega_{1}],
\end{equation}
where $w_{n}(\omega)$ is the desired signal gain at frequency $\omega$.

In some situations, passivity might be a desirable property: because
most environments are passive, interactions between a passive robot
and an arbitrary environment is guaranteed to be passive, thus
stable~\cite{slotine1991applied,albu2007unified}. Constraining the
dynamics between an exogenous input and output to be passive can be
done by constraining its frequency response.  A discrete-time transfer
function $H_{ij}(z)$ is passive if and only if
\begin{equation}
  \label{eq:6}
  \mathrm {Re}[H_{ij}(e^{j\omega T_{s})}] \geq 0, \forall \omega \in [0, \omega_{ny}].
\end{equation}
This result can be proven relatively easily following the proof of the
corresponding result for passive continuous-time LTI systems. The
above inequality is a linear parameter vector $\bm{\Theta}$.

Because both inequalities~\eqref{eq:34} and~\eqref{eq:6} are to be
satisfied over intervals, they are infinite-dimensional. For numerical
computation, one needs to approximate them as multiple point
constraints over a adequately sampled frequency grid.

\subsubsection{Disturbance rejection}
\label{sec:dist-reject}

In many robotic applications, disturbances acting on the robot are not
random white noise. Rather, they might have bounded energy, or in
other words, a finite 2-norm~\cite{doyle2013feedback}. In other cases,
they might have bounded amplitude.

Standard results in robust control theory~\cite{doyle2013feedback}
enables formulating worst-case system gains in such situations as
convex constraints on the parameter vector $\bm{\Theta}$. Consider a
signal $u_{j}[n]$, and let
$\|u_{j}[n]\|_{1}, \|u_{j}[n]\|_{2}, \|u_{j}[n]\|_{\infty}$ be its
1-norm, 2-norm (energy), $\infty$-norm (maximum amplitude)
respectively. Let $\|H_{ij}(z)\|_{2}$ be the 2-norm of the transfer
function.  The following inequalities hold:
\begin{align*}
  \|H_{ij}(z)\|_{2} =& \ \mathrm{sup} \left ( \|y_{i}[n]\|_{\infty}: \|u_{j}[n]\|_{2} \leq 1 \right ),\\
  \|h_{ij}[n]\|_{1} =& \ \mathrm{sup} \left ( \|y_{i}[n]\|_{\infty}: \|u_{j}[n]\|_{\infty} \leq 1 \right ).
\end{align*}

The above inequalities corresponds to the two examples given
earlier. One can easily show that upper bounds on $\|H_{ij}(z)\|_{2}$
and $\|h_{ij}[n]\|_{1}$ are convex in the parameter vector $\bm{\Theta}$
using the respective definitions.

\section{Experiments}
\label{sec:experiments}

\subsection{Experimental setup}
\label{sec:setup}

The experiments were performed on a 6-axis Denso VS-060 robot,
position-controlled at $\SI{125}{Hz}$. The joint position control
dynamics were experimentally identified to be a first-order LTI SISO
with time constant $\SI{0.0437}{s}$ and time delay of $\SI{36}{ms}$.
To measure contact forces, we used an ATI Force/Torque sensor on the
robot wrist at $\SI{125}{Hz}$. A low-pass filter with cut-off
frequency $\SI{73}{Hz}$ was used to prevent aliasing.

We used a personal computer running Ubuntu 16.04 with a
fully-preemptible kernel as the controller. Connections to and from
the robot and the sensor were via Ethernet TCP/IP.

In both experiments, we employed translation-only, decoupled,
Cartesian force controllers. First, wrench measurements were projected
to the global work-space coordinate frame. The projected force
components were then fed to three decoupled controllers (in X, Y and
Z). Each controller thus received a measured force and a setpoint, and
output a Cartesian position command. The three position commands were
fed to a differential IK module, which computed reference joint
positions for the robot using a Quadratic Programming solver.

Optimization programs for controller synthesis were solved with
\texttt{mosek}~\cite{mosek}. The time taken for controller synthesis
was only a few seconds. During execution, the time taken to compute
the controls was a few microseconds at each control step.


\subsection{Experiment 1: Robot hand guiding}
\label{sec:experiment}

\subsubsection{Task description}
\label{sec:task-description-2}

We instructed the operator to hold the robot end-effector and guide it
along a straight line across a distance of $\SI{60}{cm}$ along the
Y-axis of the global coordinate frame in about $\SI{3}{sec}$, see top
plot of Fig.~\ref{fig:overview}. Timing was done with a simple visual
cue. This task evaluates the effort required to teach the robot by
hand-guiding.


We implemented three controllers for admittance control in the Y
direction:
\begin{itemize}
\item \texttt{CCSa} (\underline{C}ontroller obtained by
  \underline{C}onvex \underline{S}ynthesis for \underline{a}dmittance
  control) was synthesized subject to two main specifications. First,
  the closed-loop dynamics should be stable for all environment
  stiffness up to $\SI{500}{N/mm}$. Second, the time-domain behavior
  should imitate that of a mass/spring/damper system with
  $m=\SI{1.2}{kg}, b=\SI{8}{Ns/m}, k=\SI{0}{N/m}$. The nominal human
  stiffness was modeled to be $\SI{20}{N/mm}$.

\item \texttt{CLa1} and \texttt{CLa2} (\underline{CL}assical
  \underline{a}dmittance controller 1 and 2) were designed using the
  common admittance control architecture~\cite{siciliano2016springer},
  in which the controller acts as an inverse admittance
  model. Controller \texttt{CLa1} has the same admittance parameters
  as the desired values used for CCS. This is the ideal
  dynamics. \texttt{CLa2} has the following admittance parameters:
  $m=\SI{6}{kg}, b=\SI{23}{Ns/m}, k=\SI{0}{N/m}$. The greater mass and
  damping coefficients were chosen to achieve a higher stability.

\end{itemize}
\subsubsection{Results}

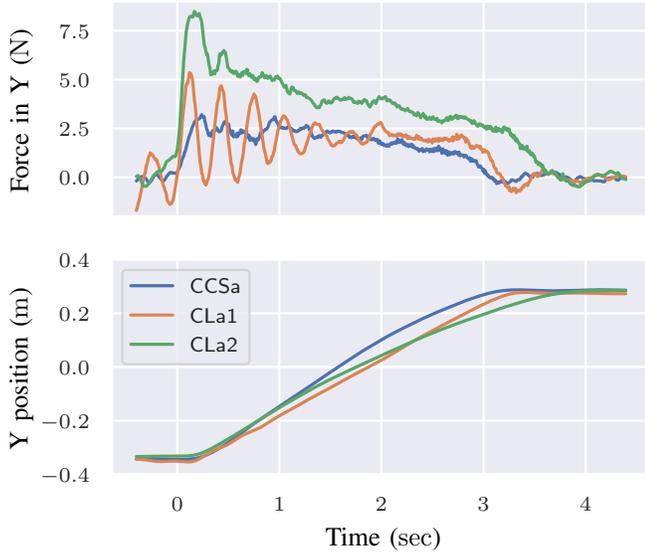
\begin{figure}[htp]
  \tikzstyle{textcube} = [color=black, fill=white]
  \centering
  \begin{tikzpicture}
    \node[anchor=south west,inner sep=0] (image) at (0,0)
    {\includegraphics[]{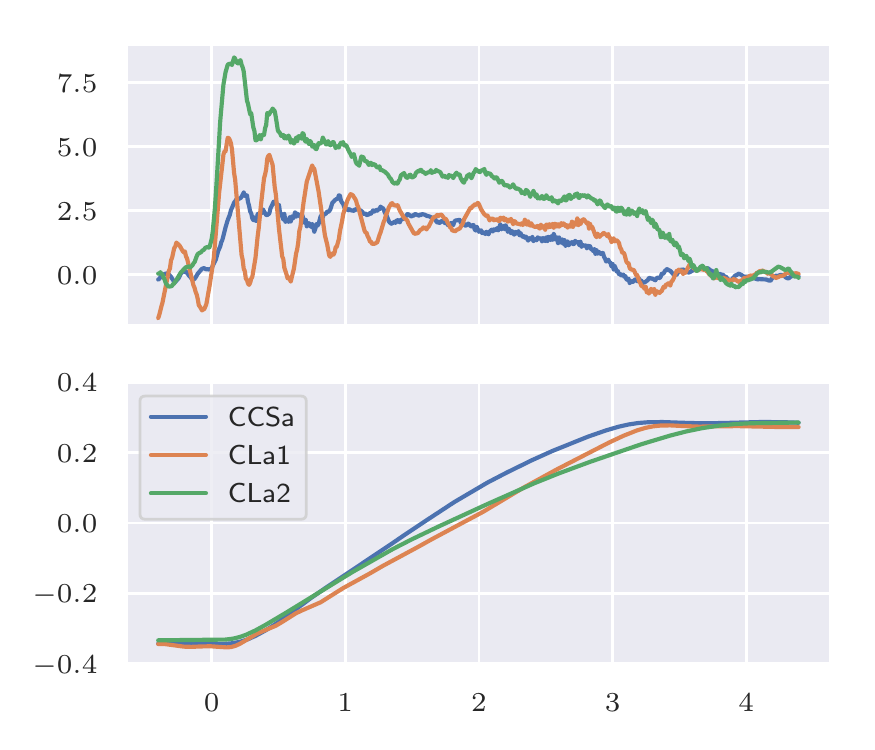}};
    \begin{scope}[x={(image.south east)},y={(image.north west)}]

      \node[anchor=north] at (0.55, 0.04) {Time ($\SI{}{sec}$)};
      \node[rotate=90] at (0.01, 0.75) {Force in Y  ($\SI{}{N}$)};
      \node[rotate=90] at (0.01, 0.3) {Y position ($\SI{}{m}$)};
    \end{scope}
  \end{tikzpicture}
  \caption{\label{fig:teaching} Robot hand guiding: operator holds the
    robot end-effector and moves it in free space over 60 cm along the
    Y-axis, within about 3 sec. Top: force measured in the Y direction
    for \texttt{CCSa} (blue), \texttt{CLa1} (orange), and
    \texttt{CLa2} (green). Bottom: Y position of the end-effector.
    Note that, under \texttt{CCSa}, the interaction was stable and
    required less effort. Under \texttt{CLa1}, the interaction was
    unstable, while under \texttt{CLa2}, it required significantly
    more effort. Video available at
    \url{https://youtu.be/uqYXVB5Sqlg}.}
\end{figure}

Fig.~\ref{fig:teaching} shows the forces and displacements in the Y
direction. One can observe that \texttt{CCSa} yields a stable behavior
and low interaction forces (less effort is required from the
operator). By contrast, under \texttt{CLa1}, a similar interaction
forces were observed, but the robot was strongly oscillatory; while
under \texttt{CLa2}, the interaction forces were significantly higher.

\label{sec:results-1}

\subsection{Experiment 2: Sliding on surfaces with different stiffnesses}
\label{sec:exper-eval-force}

\subsubsection{Task description}
\label{sec:task-description-1}

We designed a ``terraced surface'' consisting of multiple horizontal
patches with different stiffnesses and decreasing elevations, see
bottom plot of Fig.~\ref{fig:overview}. The stiffnesses and elevations
of the areas are given in Table~\ref{tab:stiffelev}. Initially the
robot was in contact with the foam patch and kept a $\SI{5}{N}$
contact force in the Z-axis. We then commanded the robot to move
uniformly along the Y-axis at $\SI{5}{mm/s}$ while maintaining the
$\SI{5}{N}$ vertical contact force.

\begin{table}[htp]
  \caption{\label{tab:stiffelev} Stiffnesses and elevations of the
    different surfaces}
  \centering
  \footnotesize
  \begin{tabular}{lrr}
    \toprule
     & Stiffness ($\SI{}{N/mm}$) & Elevation ($\SI{}{mm}$)\\
    \midrule
     foam & 3 & 0\\
     hard paper & 20 & -6.5 \\
     steel & 80 & -10\\
     aluminum & 60 & -18 \\
     aluminum & 60 & -30 \\
     carton & 4 & -35 \\
    \bottomrule
  \end{tabular}
\end{table}

We implemented three controllers for direct force control in the Z
direction:
\begin{itemize}
\item \texttt{CCSf} (\underline{C}ontroller obtained by
  \underline{C}onvex \underline{S}ynthesis for direct
  \underline{f}orce control) was synthesized subject to two main
  specifications. First, the closed-loop dynamics should be stable for
  all environment stiffnesses up to $\SI{100}{N/mm}$. Second, the
  time-domain response to step input should be that of a first-order
  system with time constant $\SI{0.17}{sec}$ at the nominal stiffness
  (foam, $\SI{3}{N/mm}$).
\item \texttt{CLf1} and \texttt{CLf2} (\underline{CL}assical direct
  \underline{f}orce controller 1 and 2) are classical
  Proportional-Integral controllers.  \texttt{CLf1} was tuned to
  achieve the same response as \texttt{CCSf} at the nominal stiffness,
  while \texttt{CLf2} was tuned to ensure stability at the highest
  expected stiffness level (steel, $\SI{80}{N/mm}$).
\end{itemize}

\subsubsection{Results}
\label{sec:results-2}

\begin{figure}[ht]
  \tikzstyle{textcube} = [color=black, fill=white]
  \centering
  \begin{tikzpicture}
    \node[anchor=south west,inner sep=0] (image) at (0,0)
    {\includegraphics[]{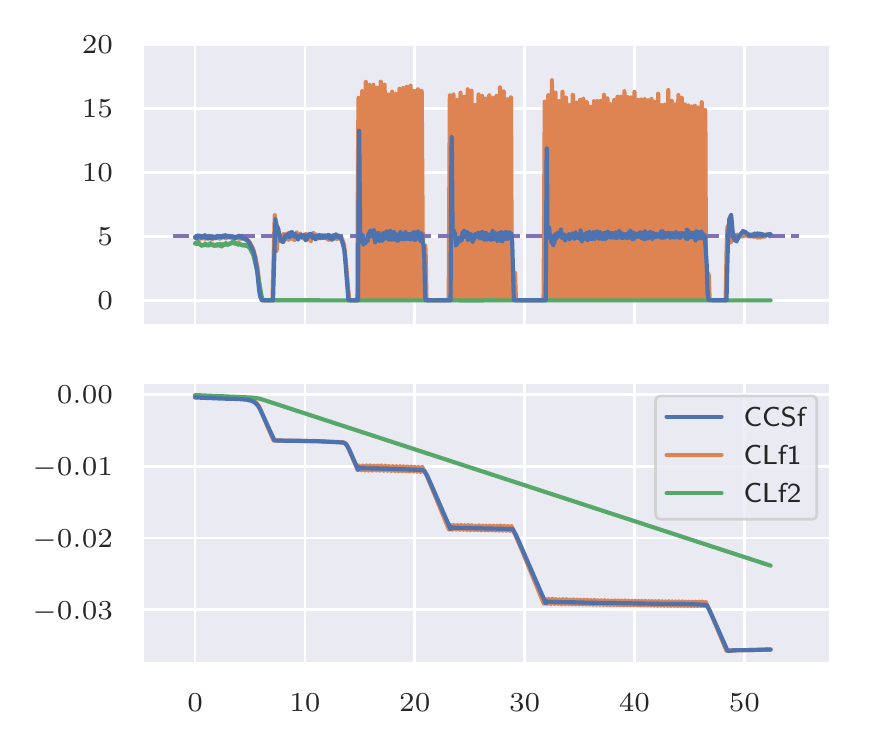}};
    \begin{scope}[x={(image.south east)},y={(image.north west)}]

      \node[anchor=north] at (0.55, 0.04) {Time ($\SI{}{sec}$)};
      \node[rotate=90] at (0., 0.75) {Force in Z ($\SI{}{N}$)};
      \node[rotate=90] at (0., 0.3) {Z position ($\SI{}{m}$)};
    \end{scope}
  \end{tikzpicture}
  \caption{\label{fig:force} Sliding on surfaces with different
    unknown stiffnesses. The desired contact force (in Z) was
    $\SI{5}{N}$, while the desired velocity in the Y direction was
    $\SI{5}{mm/s}$. Top: force measured in the Z direction. Bottom: Z
    position of the end-effector. Under \texttt{CCSf}, the robot could
    maintain contact at $\SI{5}{N}$ on all surfaces and showed no
    vibrations. Under \texttt{CLf1}, instability occurred when the
    tool slid over the harder surfaces (steel, aluminum). Under
    \texttt{CLf2}, the robot was too sluggish to follow the changes in
    surface elevation, losing contact immediately. Note that, for
    \texttt{CCSf} and \texttt{CLf1}, at the transitions between two
    surfaces, the end-effector temporarily loses contact with the
    surfaces, leading to zero contact force.  Video available at
    \url{https://youtu.be/uqYXVB5Sqlg}.}
\end{figure}
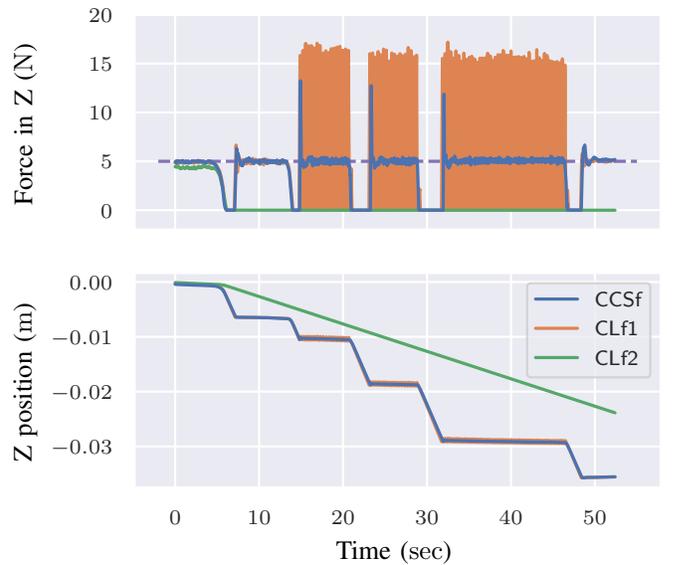

Fig.~\ref{fig:force} shows the forces and displacements in the Z
direction. Under \texttt{CCSf}, the robot maintained contact at
$\SI{5}{N}$ during the whole experiment, regardless of material,
except at brief transition periods between surfaces. The recorded
contact force had sharp overshoots when making contact with the stiff
materials, but showed no noticeable vibrations.

Under \texttt{CLf1}, the force tracking loop was stable only on foam
and paper, and became unstable on stiffer materials such as steel and
aluminum, as evident from the strong oscillations in the force
measurements. Under \texttt{CLf2}, the robot response became so slow
that it was in fact unable to track the changes in surface elevation.

\subsection{Discussion}
\label{sec:discussion}

{\small

The experimental results demonstrate that CCS controllers perform
significantly better than their classical counterparts. The robot
remained stable when in contact with environments up to 27 times
stiffer than the nominal value. In addition to being robustly stable,
there were no visible loss in nominal performance as compared to
hand-tuned Admittance and PI controllers.

One main reason for this superior performance is that classical
controllers have fixed structures, hence are limited by the number of
tunable parameters. By contrast, CCS controllers were optimized in the
space of \emph{all stabilizing controllers}: they can therefore
achieve a significantly higher level of performance.


In the limit, there will however be an unavoidable trade-off between
nominal performance and robust stability. For instance, if one wishes
to synthesize a CCS controller for robot hand guiding (\texttt{CCSa})
that is robustly stable against environments stiffer than
$\SI{500}{N/m}$, then one will have to sacrifice somehow the
responsiveness of the robot. Such a trade-off has also been reported
in other contexts~\cite{balas1990robustness,boyd1990linear}.

}

\section{Conclusion}
\label{sec:conclusion}

{\small

  We have proposed a new approach for synthesizing controllers for
  contact. Our CCS controllers are robustly stable \emph{and} achieve
  high performance. Compared to approaches reviewed in
  Section~\ref{sec:related-work}, CCS can account for most relevant
  uncertainties including time-delays, time-discretization, high-order
  dynamics, while remaining stable against an unprecedentedly wide
  range of environment stiffnesses (up to 27 times).

  Synthesizing controllers numerically has two operational
  benefits. First, one can implement a synthesized controller in a
  simple fashion, requiring no complex online computation. Second,
  numerical synthesis is efficient thanks to available powerful convex
  optimization solvers, reducing the need for manual hand tuning. This
  is especially relevant for complex systems and/or complex
  performance specifications.



  One assumption we make in this paper is that the dynamics along the
  three translation axes are decoupled: extending CCS to rotational
  motions and coupled dynamics is an important research
  direction. This requires extending the proposed method to handle
  Multiple-Input Multiple-Output (MIMO) systems. A particular
  difficulty we foresee will be to derive convex robust stability
  conditions for such systems.

  Since CCS is particularly adapted to handle fast switching contacts
  between the robot and environments with widely varying and unknown
  stiffnesses, applying CCS to legged robots is another promising
  avenue for future research.

}


\subsection*{Acknowledgments}

This work was partially supported by A*STAR, Singapore, under the AME
Individual Research Grant 2017 (Project A1883c0008).

\bibliographystyle{IEEEtran}
\bibliography{library}

\end{document}